\documentclass[11pt]{article}

\usepackage[final]{acl}

\usepackage{times}
\usepackage{latexsym}

\usepackage[T1]{fontenc}

\usepackage[utf8]{inputenc}

\usepackage{microtype}

\usepackage{inconsolata}

\usepackage{graphicx}

\usepackage{lipsum}
\usepackage{verbatim}
\usepackage{listings}
\usepackage{xcolor}
\usepackage{booktabs}
\usepackage{multirow}
\usepackage{array}
\usepackage{subcaption}
\usepackage{fontawesome}
\usepackage{amsmath}
\usepackage{longtable}
\usepackage{xurl}
\usepackage{supertabular}  

\title{The Course of News Events: A Comparison of Bottom-Up and Top-Down Approaches for Collecting Text-Based Data about Disasters}

\author{\\
	\textbf{Brielen Madureira\textsuperscript{1,2}},
	\textbf{Andreas Niekler\textsuperscript{1,3}},
	\textbf{Mariana Madruga de Brito\textsuperscript{2,1}}
	\\\\
	\textsuperscript{1}\small LeipzigLab -- Climate Discourse, Leipzig University, Germany \hfill
	\textsuperscript{2}\small Helmholtz Centre for Environmental Research, Germany \\
	\textsuperscript{3}\small Computational Humanities, Leipzig University, Germany \hfill
	\\
	\small{
		\textbf{Correspondence:} \href{mailto:brielen.madureira@uni-leizig.de}{brielen.madureira@uni-leizig.de}
	}
}

\begin{document}
\maketitle
\begin{abstract}
	News articles are an important source of information on disaster impacts and adaptation. A key methodological challenge in socio-environmental studies is how to select a representative data sample. Two approaches are common: querying news databases top-down with the aid of an existing disaster inventory or using NLP methods to cluster news texts bottom-up based on temporal and spatial features. Using a dataset of German news about landslides worldwide, we compare these approaches and discuss variations in event coverage. Such research design decision can influence the resulting news sample, affecting its use in studies of inequality in media coverage, disaster monitoring and inventory enrichment.
\end{abstract}

\section{Introduction}
Understanding how humans experience and respond to disasters calls for interdisciplinary collaboration between environmental and (computational) social science \citep{meehl-2000,albeverio2006extreme,mcphillips-2018,de2023computational}. At this intersection blossom fundamental questions about the consequences of climate hazards to society, the relation between adaptation measures and risk reduction, and inequalities in exposure and impact across societal groups. Ensuing findings can ultimately inform allocation of disaster-relief funds \citep{Chapman2022} and democratic decision making \citep{soroka2012mass}.

Synthesizing such knowledge requires gathering information that sprouts on various sources, from sensor-based measurements to digital documents. When text data is in view, the Natural Language Processing field joins that interdisciplinary table with methods for e.g.~classification, information extraction, geoparsing and modelling complex social constructs in language use \citep{debrito2026assessing}. 

\begin{figure}[t]
	\centering
	\includegraphics[trim={0 3.5cm 0 0},clip,width=\columnwidth]{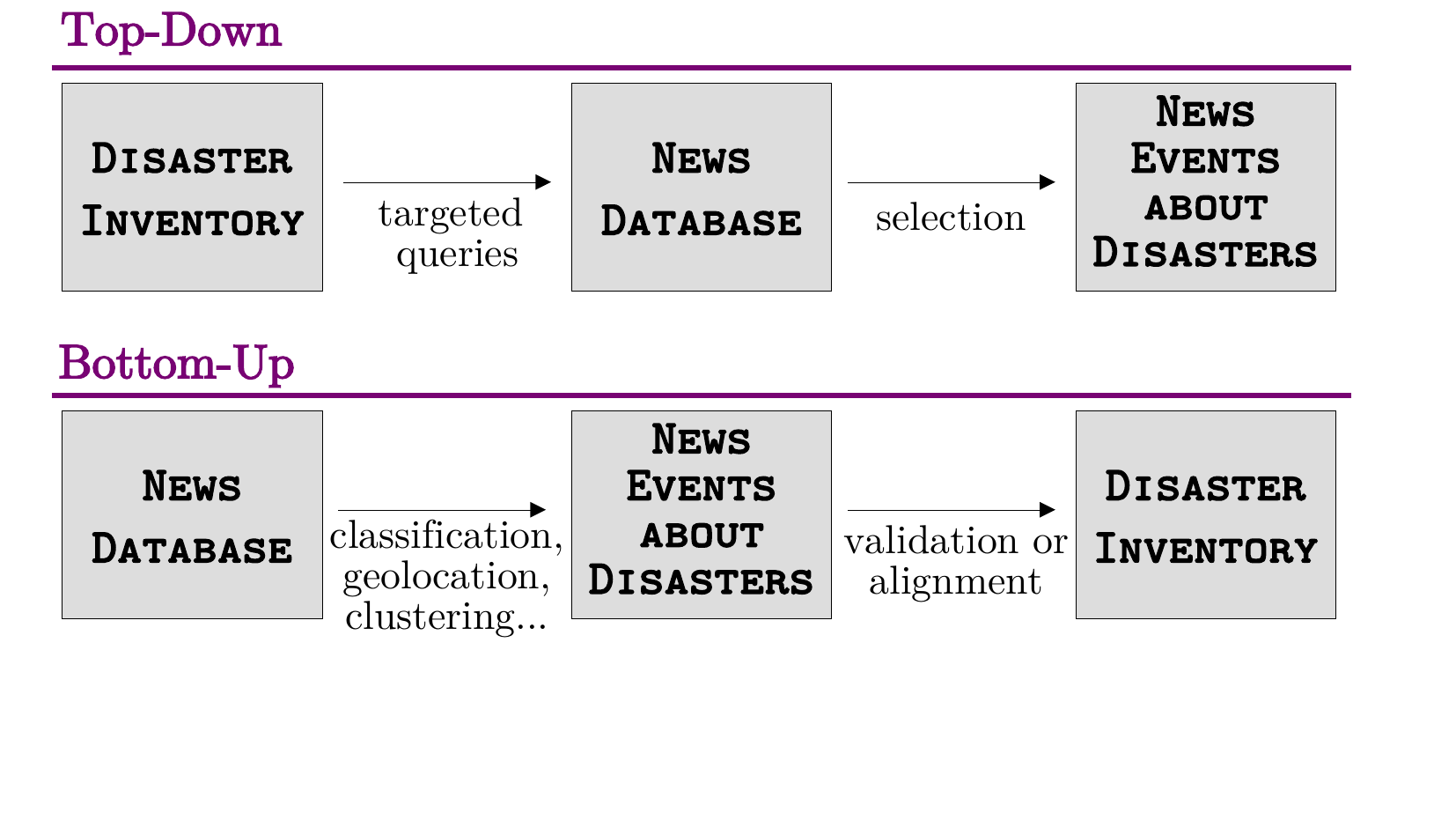}
	\caption{Illustrative comparison of two approaches to detect references to disasters in the news.}
	\label{fig:comparison}
\end{figure}

A fundamental challenge comprises identifying the time, location and impact of disasters 
(e.g.~landslides, wildfires and floods). Numerous studies (see Section \ref{sec:lit}) use global disaster inventories like EM-DAT \citep{Delforge2025}, which is based on high-quality, manually curated data but is also incomplete and unavoidably biased. As a consequence, when the scientific community overly relies on a single source serving as ground truth, the derived collective knowledge may overfit traits of the database rather than the actual phenomena.

To counteract this problem, news databases can provide additional event information via two approaches (Figure \ref{fig:comparison}): \textbf{top-down} procedures use known events in external disaster inventories to \textbf{query} news databases for targeted content (e.g.~\citealp{cai_2025}), whereas \textbf{bottom-up} procedures identify, geolocate and cluster news into segmented events, which can then be \textbf{aligned} to or validated against external inventories (e.g.~\citealp{Valkenborg2026}). But neither is infallible: while the first overlooks events not recorded in inventories, the latter ignores events that were not deemed newsworthy by the media represented in the news database. 

This paper looks more closely into this matter. We compare news events identified via top-down and bottom-up approaches in dataset of German news about landslides and discuss their advantages and shortcomings. Such methodological insights, grounded in empirical observations, can strengthen NLP-supported socio-environmental studies. 


\section{Related Work}
\label{sec:lit}
The International Disaster Database (EM-DAT, \citealp{Delforge2025}) is a widely used global inventory of 27k+ events, often regarded as the ground truth. However, its coverage is constrained by its inclusion criteria and the difficulty of gathering detailed information for \textit{all} countries. Even the recorded events have missing data \citep{Jones2022}. Despite such limitations, hundreds of empirical studies rely on it \citep{Jones2023}. 

Concomitantly, news articles have long been seen as a source of information about disasters that serves to create or enrich inventories \citep[\textit{inter alia}]{Guzzetti1994,Llasat2009,Taylor2015,henrique_lima_alencar_flash_2024,Sodoge2024,Avcolu2025}, monitor hazard events in near real-time \citep{tanev2008} and study media attention to them \citep{Yan2015,Kong07022026}. Disaster inventories and impact information derived from large-scale textual data require careful validation \citep{debrito2026assessing}. For example, observations can be aligned with EM-DAT entries for calibration \citep{Li2025,dahr2026climate,Valkenborg2026}.

\section{Data and Event Matching}
We used a dataset comprised of almost 55k news articles in German about landslides worldwide, constructed by \citet{madureira2026loudrumbleshitnewsstands}. That study queried the \texttt{wiso-net} news database (in the period from 2000 to 2024) using landslide-related keywords, identified relevant articles and geoparsed them at the country level with the aid of Large Language Models (LLMs). Then, \textit{news events} were identified in a bottom-up approach. A news event was defined as a sequence of news articles referring to landslides in the same country, starting on the first day with at least one observation and ending right before at least $\theta=5$ consecutive days\footnote{This parameter can vary, but we kept it for consistency.} without any coverage occurred.

Our analysis was based on matching events between the bottom-up news events from that study and the top-down events in EM-DAT. For that, we extracted a list of 2,014 landslide events (of type main or associated) for 138 countries from EM-DAT, whose entries contain the event's onset date, country and location, among other information. A visualisation of how bottom-up events and top-down EM-DAT entries are temporally dispersed is presented in Figures \ref{fig:top_all} and \ref{fig:top_2024} (Appendix).

\begin{figure}[t]
	\centering
	\includegraphics[trim={0 6.5cm 12cm 0},clip,width=\columnwidth]{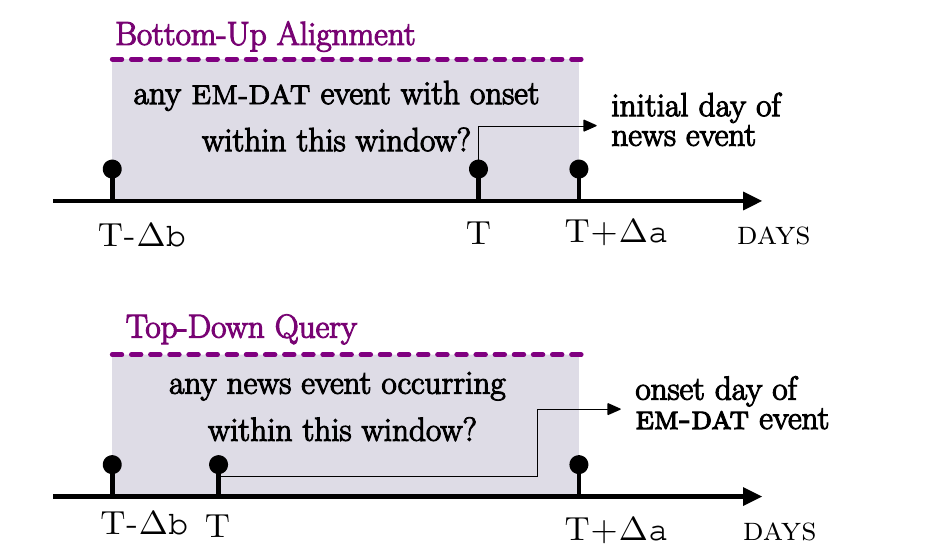}
	\caption{Illustration of the event matching procedures.}
	\label{fig:alignment}
\end{figure}

\paragraph{Bottom-up approach} Using time series segmentation upon the geolocated news articles, \citet{madureira2026loudrumbleshitnewsstands} identified 4,567 news events for 152 countries (see that publication for methodological details). For each news event, the provided data file contained metadata such as its initial date, duration in days and the associated news articles. To perform event matching, we considered a news event to be \textbf{\textit{temporally aligned}} with an EM-DAT entry if the entry's onset date is near the beginning of the news event (from $\Delta_b$ days before to $\Delta_a$ days after its initial date), as in Figure \ref{fig:alignment} (top).

\paragraph{Top-down approach} For each EM-DAT entry, event matching was performed by \textbf{\textit{querying}} the news database around the entry's onset date (i.e.~from $\Delta_b$ days before to $\Delta_a$ days after it) for news about the country. If at least one news article was retrieved, we considered that the EM-DAT event \textbf{\textit{temporally coincided}} with a news event, as in Figure \ref{fig:alignment}. Instead of setting a fixed retrieval period, we allowed for news events with varying durations following the same rationale used in the bottom-up approach: the news event started on the first day with a news article in $[\Delta_b, \Delta_a]$ and ended before $\theta$ days without any observed news for the country.

To select the $\Delta$ values for this study, we regarded the onset date in EM-DAT as correct and acknowledged that some events may take a few days to appear in international news. Therefore, for the top-down approach, we set $\Delta_b=1$ and $\Delta_a=5$, meaning that the queried news must start within 5 days after the known onset date. The 1 day before should account for potential errors and early warnings. Conversely, for the bottom-up approach, we set $\Delta_b=5$ and $\Delta_a=1$, meaning that, if an event is already in the news, its onset date cannot be later than the first news (1-day tolerance), but could have been some days earlier. Future studies can investigate the impact of using other parameters.

\section{Analysis}
\label{sec:analysis}	
We first compared the outcomes of the two event matching strategies by quantifying (i) how many bottom-up news events temporally \textbf{\textit{aligned}} with an EM-DAT entry and (ii) how many EM-DAT entries temporally matched news events through \textbf{\textit{querying}}. As shown in Table \ref{tab:results}, the bottom-up approach identified more than twice as many news events as the number of entries in EM-DAT. However, less than 17\% of the news events temporally aligned with an EM-DAT entry. In contrast, the top-down queries successfully queried temporally coinciding news events for almost 43\% of the EM-DAT entries. 

The alignment and querying procedures are not bijective. On the one hand, EM-DAT entries with distinct onset dates can end up querying the same news event at different days throughout it (especially news events of long duration). On the other hand, bottom-up news events can be aligned with multiple EM-DAT entries whose onset dates are close in time. Such partial or multiple matches between EM-DAT entries and news events would require further disambiguation steps. 

\begin{figure}[t]
	\centering
	\includegraphics[width=0.45\columnwidth]{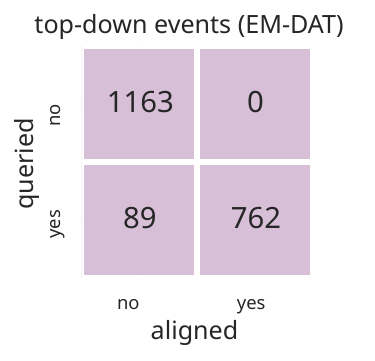}
	\includegraphics[width=0.45\columnwidth]{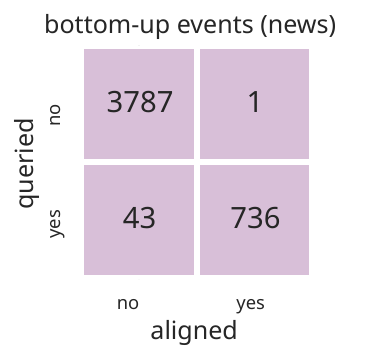}
	\caption{Confusion matrices with the number of aligned and/or queried events in both sources.}
	\label{fig:cfs}
\end{figure}

Figure \ref{fig:cfs} depicts the confusion matrices with the overlap between aligned and queried events in each type of event source. The 851 successful queries covered 779 unique news events. 89 queries matched news events midway through and 60 news events were queried by more than one EM-DAT entry. Such cases require post-processing decisions on whether two distinct news topics were inappropriately merged in the bottom-up approach (e.g.~because they overlapped in time), or whether the event matching was simply spurious in terms of content. Even in the top-down approach, such reasoning would require bottom-up information about when underlying news events begin and end.

Figure \ref{fig:cfs} also shows that 737 aligned bottom-up events covered 762 EM-DAT entries (one of them twice). 26 news events had multiple alignments with EM-DAT entries and only one news event was aligned but not detected via querying (because the query captured an immediately preceding news event). Again, post-processing decisions would be needed to determine which EM-DAT entry properly reflects the content of the news event.

\begin{table}
	\centering
	\small
	
	\begin{tabular}{ll}
		\toprule
		
		\textbf{Bottom-up} & \\
		~~~~~news events & 4,567 \\
		~~~~~aligned to EM-DAT & ~~~737 (16.1\%) \\
		\cmidrule{1-2}
		\textbf{Top-down} & \\
		~~~~~EM-DAT events & 2,014 \\
		~~~~~queried in the news & ~~~851 (42.2\%)\\
		\cline{1-2}
		\bottomrule
	\end{tabular}
	
	\caption{Number of bottom-up news events and top-down EM-DAT entries with the portion of temporally matched events.}
	\label{tab:results}
\end{table}

The core finding is that the two approaches coincided in temporally matching 762 EM-DAT entries to 736 news events. 57.7\%~of EM-DAT events had empty queries, meaning that there was no relevant news near their onset dates. Similarly, 82.9\% of the bottom-up news events did not align with any EM-DAT entry. These results indicate that, on the one hand, EM-DAT contains entries that could not be detected in the German news, and, on the other hand, that the media may form news events that are not recorded in EM-DAT. Had only top-down queries been used, many news events would have been missed, while a bottom-up approach alone would have ignored more than half of the known landslides worldwide that did not appear in the German media (as represented in this data sample).

\begin{figure}[t]
	\centering
	\includegraphics[width=\columnwidth]{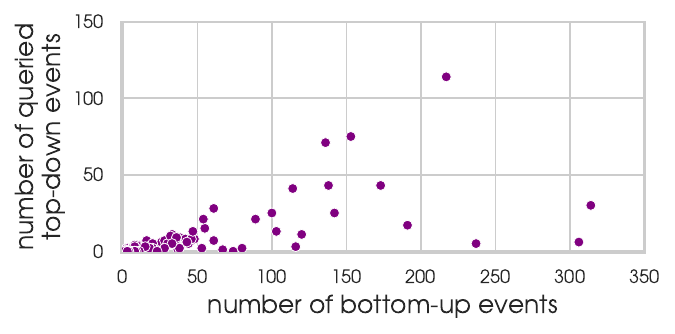}
	\caption{Number of detected news events by country.}
	\label{fig:n-events}
\end{figure}

Although alignment can be used to calibrate bottom-up news events, in practice the decision about which news sample to use lies between the \textit{queried} EM-DAT entries or \textit{all} bottom-up news events. Country-level analyses would be impacted by this research design choice. Figure \ref{fig:n-events} shows how many events were detected for observed countries in each approach. The maps in Figures \ref{fig:map-emdat}, \ref{fig:map-queried} and \ref{fig:map-news} (Appendix) illustrate the spatial coverage and gaps of each approach. The bottom-up approach generally detected more news events per country than the successful queries, with noticeable variations in the resulting sample distribution. For instance, 54.3\% of the bottom-up news events are assigned to the Global South, compared to 81.4\% of the queried news events; 47.7\% of the bottom-up news events refer to high income countries \textit{versus} 20.6\% of the queried news events; and 35.8\% of the former are in Europe in contrast to 9.7\% of the latter (see detailed Tables in the Appendix). 

\section{Discussion}
Numerical results alone may give the impression that bottom-up news events are more advantageous, as they exceed the number of queried news events and can also aid near real-time monitoring. But caution is warranted: news events do not always reflect new real-world events relevant to disaster inventories. False positives stemming from errors in the NLP pipeline add noise to the sample and can form spurious news events. Moreover, many news events bring the \textit{topic} of landslides in a country to the public's attention without reporting about recent or ongoing events. Based on an initial manual verification of 150+ news events, we identified a few common types of bottom-up news events:

\begin{itemize}
	\setlength\itemsep{-0.2em}
	
	\item \textbf{concrete in progress}: texts that refer to ongoing or recent specific landslides in a country;
	
	\item \textbf{concrete but past}: texts that refer to landslides that happened in the near or distant past in a country (e.g.~aftermath of an event after some time, memories of previous disasters, judicial decisions about past events, pre-historic events that shaped landscapes etc.);
	
	\item \textbf{indefinite}: texts about landslides in a country with no reference to specific events having taken place (e.g.~studies, aggregated outcome during a period, warnings, references to risk, preventive measures, hypothetical events, the phenomenon as a whole, underspecified or implicit sentences etc.); 

	\item \textbf{vague}: texts that refer to landslides in fiction works or just as rhetorical examples;
	
	\item \textbf{fully false positives}: misclassified texts (e.g.~figurative uses of landslide terms, wrong or unclear geolocation).
	
\end{itemize}

News events detected bottom-up can also mix texts of different types that coincide in time or repeat the same text in distinct events when the news article is republished after many days.

Both approaches have advantages and disadvantages and should be regarded as complementary. The bottom-up approach may detect events not present in EM-DAT, but it also captures news on the general \textit{topic} of landslides in a country and is prone to noise due to the impossibility of perfect classification and geolocation. It is suitable for studies of media attention but requires refinement to serve as a source of information to create inventories. The top-down approach is more controllable but requires disambiguation of news events captured midway through and remains constrained to a subsample of known events. It may be more appropriate for enriching existing disaster databases with impact data, but it introduces the inventory's biases into subsequent conclusions about media coverage. Both approaches share the limitation that events not covered by the media represented in the sample are left out, making a top-down reference important for quantifying the lack of media coverage. 

The biases in media coverage and EM-DAT inclusion are arguably not random. EM-DAT focuses on severe events, whereas the media may devote more attention to events in more populous or economically relevant regions, or to events that attract greater public interest. Further studies can unravel the differences in which happenings get to be considered ``events'' in each source.

\section{Conclusion}
Despite their potential for broad temporal and geographic coverage of disaster events, news-based datasets are shaped by how the media selects, frames and reports disasters. Besides, articles often refer to past or hypothetical events, and unstructured information is sometimes left underspecified. 

Large-scale datasets can conceal shortcomings that affect derived conclusions. Automated information extraction from texts does not eliminate all false positives, thereby hindering temporal and spatial precision. Regardless of whether a sample of disaster news is constructed via a bottom-up or top-down approach, measures to ensure data quality and to understand which events were (or were not) captured are imperative.  Diagnostic analyses can expose problems that otherwise go unnoticed in the big data paradigm. By discussing observed differences between the two approaches, this work has contributed to best practices in text-based climate impact research, supporting authors in making informed methodological choices.


\bibliography{custom}

\appendix

\section{Appendix}
\subsection*{Limitations}
This analysis faces some limitations. First, it relies on country-level event matching, which may merge distinct landslides occurring close in time within the same country. 

Important parameters ($\theta$, $\Delta_a$ and $\Delta_b$) were set to fixed values. Further sensitivity analyses should examine how other values would affect the results and whether different countries or types of events require different values (e.g.~events in a neighbouring country may possibly hit the headlines faster than events in geographically and culturally distant countries). 

Depending on the use case, further investigation is also needed to refine the bottom-up approach so as to capture only concrete and current real-world events. 

Finally, both the alignment and the querying event matching strategies were limited to the \textit{temporal} aspect and the country. Detailed error analyses regarding \textit{content} matching should still be conducted to ensure that the content of the news event actually refers to the EM-DAT entry. While the specific location toponyms in EM-DAT could help further refine the queries, news articles do not always mention them, typos and name variation are frequent, and specific places may remain unreported when hazards affect a broad region. 

\subsection*{Details and Additional Results}
Table \ref{tab:dists} shows the distribution of identified and queried news events by country category, mentioned in Section \ref{sec:analysis}. Table 3 lists the number of (aligned) news events and (queried) EM-DAT entries by country. 

\begin{table}[h]
	\centering
	\small
	
	\begin{tabular}{lrr}
		\toprule
		& \multicolumn{2}{c}{\textbf{\% of news events}} \\
		\cmidrule{2-3}
		& \textbf{Identified} & \textbf{Queried} \\
		& \textbf{bottom-up} & \textbf{ top-down} \\
		\cmidrule{1-3}
		\textbf{Development Group} &  &  \\
		~~~~~Global North & 45.63 & 18.56 \\
		~~~~~Global South & 54.36 & 81.43 \\
		\cmidrule{1-3}
		
		\textbf{Income level} &  &  \\
		~~~~~High & 47.79 & 20.68 \\
		~~~~~Upper middle & 24.72 & 43.12 \\
		~~~~~Unknown & 1.05 & 0.94 \\
		~~~~~Lower middle & 21.94 & 30.55 \\
		~~~~~Low & 4.48 & 4.70 \\
		\cmidrule{1-3}
		
		\textbf{Region} &  &  \\
		~~~~~Africa & 6.28 & 5.64 \\
		~~~~~Americas & 23.42 & 24.44 \\
		~~~~~Asia & 31.74 & 58.04 \\
		~~~~~Europe & 35.88 & 9.75 \\
		~~~~~Oceania & 2.64 & 2.11 \\
		
		\cline{1-3}
		\bottomrule
	\end{tabular}
	
	\caption{Distribution of bottom-up news events and top-down queried news events by country category.}
	\label{tab:dists}
\end{table}

\begin{figure*}[t]
	\centering
	\includegraphics[width=\textwidth]{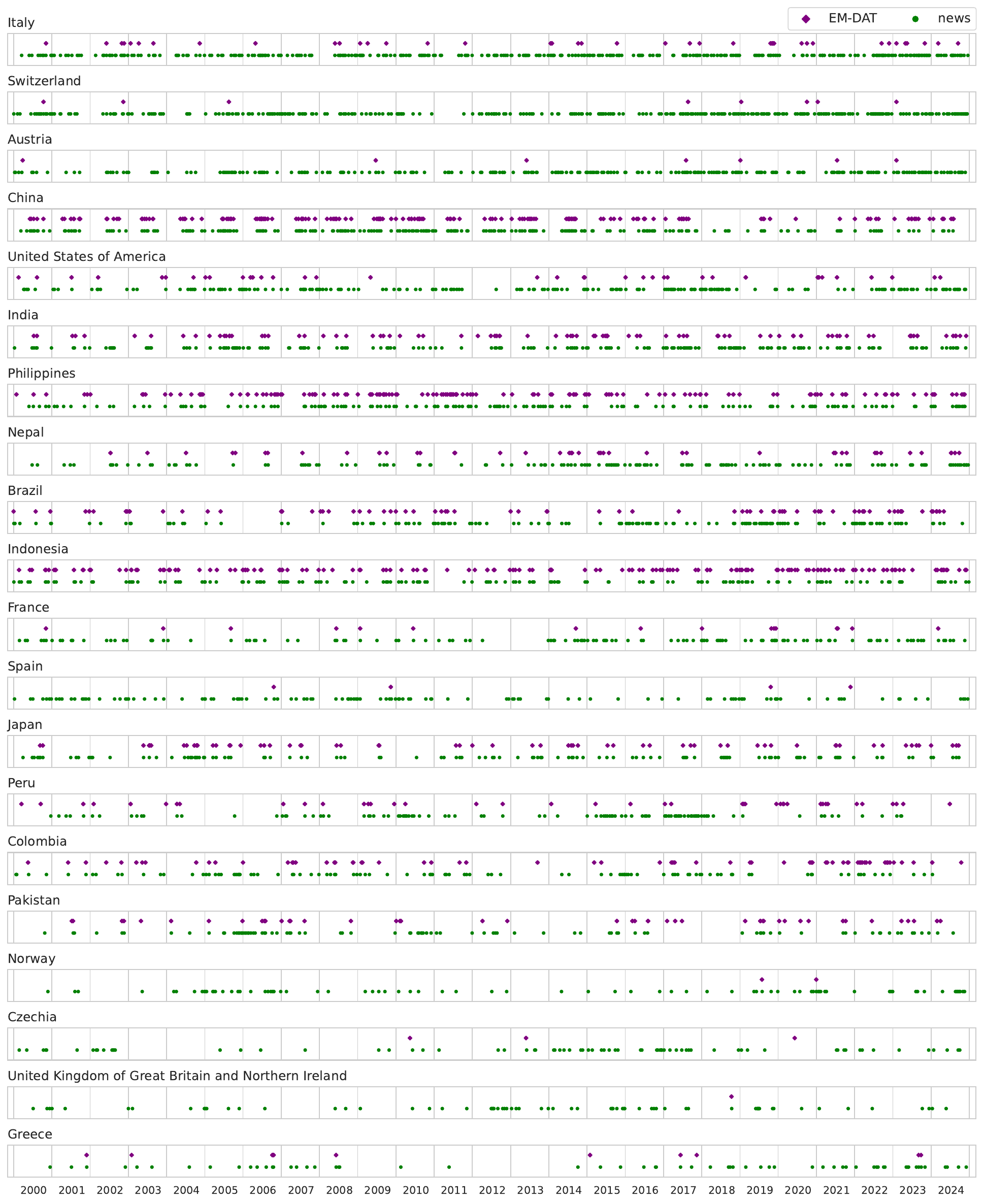}
	\caption{Broad overview of the temporal dispersion of the onset days of EM-DAT entries (green circles) and initial days of news events (purple diamonds) for the 20 countries with most news events. The x-axis lists all days in the period from Jan 1, 2000 to Dec 31, 2024.}
	\label{fig:top_all}
\end{figure*}

\begin{figure*}[t]
	\centering
	\includegraphics[width=\textwidth]{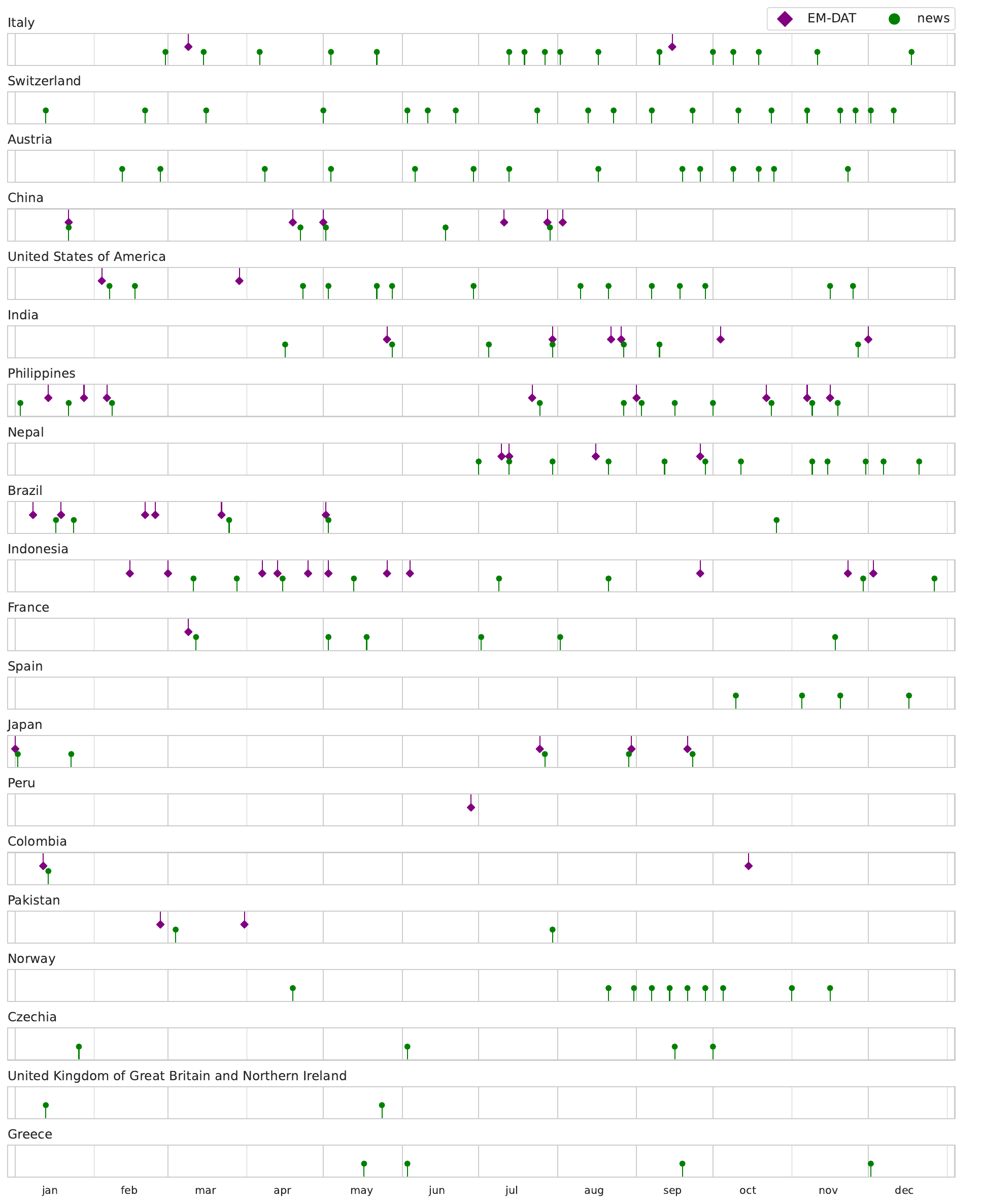}
	\caption{Broad overview of the temporal dispersion of the onset days of EM-DAT entries (green circles) and initial days of news events (purple diamonds) for the 20 countries with most news events. The x-axis lists all days in the period from Jan 1, 2024 to Dec 31, 2024.}
	\label{fig:top_2024}
\end{figure*}

Figures \ref{fig:top_all} and \ref{fig:top_2024} plot the onset date of EM-DAT events and the first active day of news events along the time period (x axis) for the 20 countries with the most observed news events. Figure \ref{fig:top_all} is a broad overview of the whole period, whereas Figure \ref{fig:top_2024} zooms in on the year 2024 for a more detailed inspection. These illustrations show the three typical (mis)matching cases: isolated EM-DAT events that (most likely) had no media coverage in German newspapers; isolated news events that do not seem to have been recorded in EM-DAT; and the matches, i.e.~EM-DAT and news events that occur in neighbouring days, subject to alignment and querying.

\begin{figure*}[t]
	\centering
	\fbox{\includegraphics[width=\textwidth]{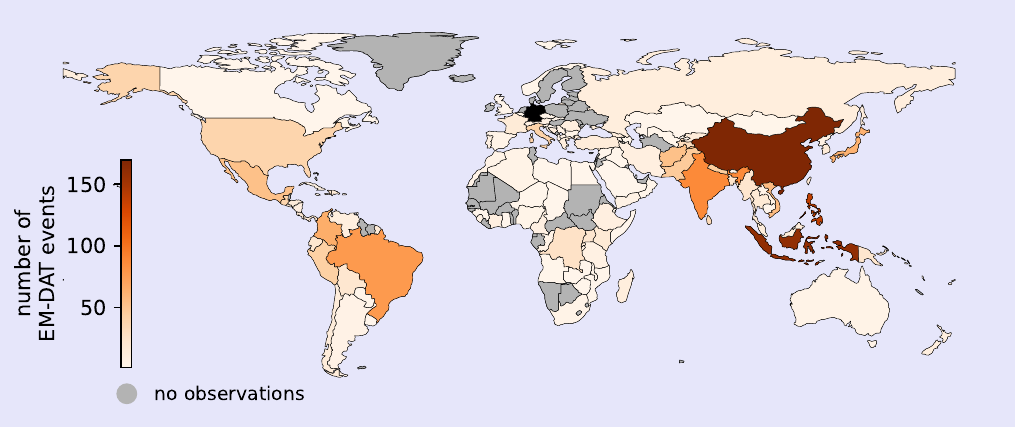}}
	\caption{Spatial distribution of EM-DAT entries referring to landslides. Germany (in black) was not analysed.}
	\label{fig:map-emdat}
\end{figure*}

\begin{figure*}[t]
	\centering
	\fbox{\includegraphics[width=\textwidth]{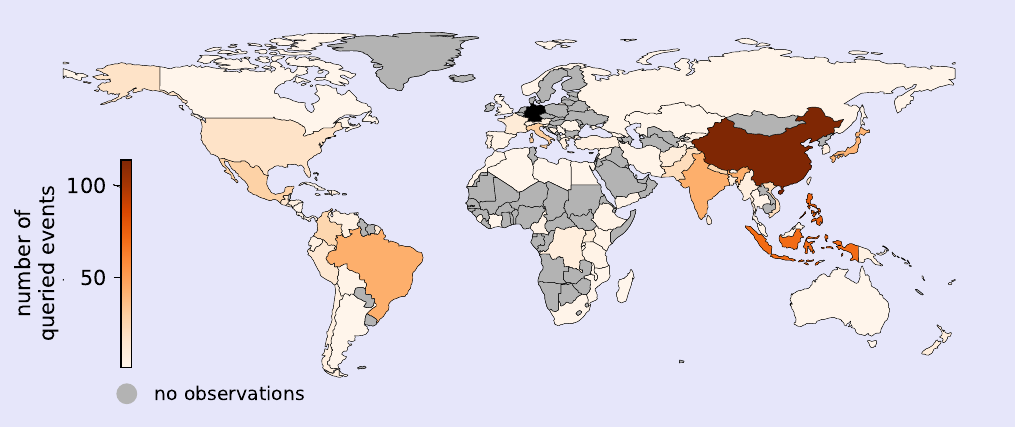}}
	\caption{Spatial distribution of EM-DAT entries referring to landslides that could be queried (top-down) in the news database. Germany (in black) was not analysed.}
	\label{fig:map-queried}
\end{figure*}

\begin{figure*}[t]
	\centering
	\fbox{\includegraphics[width=\textwidth]{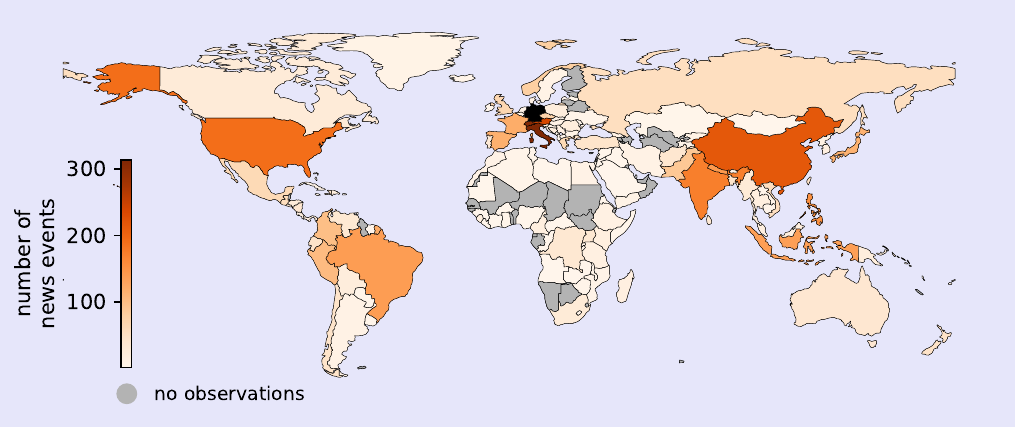}}
	\caption{Spatial distribution of (bottom-up) news events referring to landslides. Germany (in black) was not analysed.}
	\label{fig:map-news}
\end{figure*}

Figures \ref{fig:map-emdat}, \ref{fig:map-queried} and \ref{fig:map-news} are meant for comparing the differences in spatial distribution of each source: EM-DAT events, EM-DAT events for which news articles could be queried, and news events. This evidence indicates that the chosen method can yield samples with considerable differences in coverage that would influence subsequent analyses.

\clearpage

\footnotesize
\tablehead{%
	\toprule
	& \multicolumn{2}{c}{\textbf{Bottom Up}} & \multicolumn{2}{c}{\textbf{Top Down}} \\ \cmidrule(lr){2-3}  \cmidrule(lr){4-5}
	& news events & aligned & EM-DAT & queried \\
	\cmidrule(lr){2-3}\cmidrule(lr){4-5}}

\tabletail{%
	\bottomrule \\
	& & & \multicolumn{2}{r}{continues...} \\
	}

\tablelasttail{\bottomrule}
	
\bottomcaption{Number of (aligned and total) news events and (queried and total) EM-DAT entries by country.}

\begin{supertabular}{lrrrr}
	ABW & 1 & 0 & 0 & 0 \\
	AFG & 38 & 8 & 53 & 9 \\
	AGO & 1 & 0 & 2 & 0 \\
	ALB & 2 & 0 & 4 & 0 \\
	ARG & 10 & 1 & 5 & 1 \\
	ARM & 0 & 0 & 1 & 0 \\
	ASM & 0 & 0 & 1 & 0 \\
	AUS & 37 & 1 & 4 & 1 \\
	AUT & 237 & 4 & 7 & 5 \\
	AZE & 0 & 0 & 2 & 0 \\
	BDI & 7 & 2 & 8 & 2 \\
	BEL & 6 & 0 & 2 & 0 \\
	BFA & 2 & 0 & 0 & 0 \\
	BGD & 42 & 8 & 23 & 8 \\
	BGR & 5 & 0 & 3 & 0 \\
	BHS & 1 & 0 & 0 & 0 \\
	BIH & 17 & 5 & 10 & 5 \\
	BOL & 22 & 3 & 21 & 3 \\
	BRA & 138 & 40 & 77 & 43 \\
	BRB & 0 & 0 & 1 & 0 \\
	BTN & 5 & 1 & 2 & 1 \\
	CAF & 1 & 0 & 0 & 0 \\
	CAN & 27 & 1 & 1 & 1 \\
	CHE & 306 & 6 & 8 & 6 \\
	CHL & 45 & 6 & 14 & 6 \\
	CHN & 217 & 80 & 170 & 114 \\
	CIV & 2 & 1 & 9 & 1 \\
	CMR & 18 & 4 & 8 & 4 \\
	COD & 36 & 8 & 26 & 8 \\
	COG & 1 & 0 & 4 & 0 \\
	COL & 100 & 23 & 65 & 25 \\
	COM & 1 & 0 & 2 & 0 \\
	CPV & 0 & 0 & 1 & 0 \\
	CRI & 16 & 3 & 18 & 3 \\
	CUB & 6 & 2 & 7 & 2 \\
	CYM & 2 & 0 & 0 & 0 \\
	CZE & 74 & 0 & 3 & 0 \\
	DJI & 0 & 0 & 2 & 0 \\
	DMA & 7 & 2 & 2 & 2 \\
	DNK & 11 & 0 & 0 & 0 \\
	DOM & 17 & 6 & 16 & 6 \\
	DZA & 4 & 1 & 4 & 1 \\
	ECU & 44 & 5 & 21 & 5 \\
	EGY & 9 & 1 & 1 & 1 \\
	ERI & 1 & 0 & 0 & 0 \\
	ESP & 116 & 3 & 4 & 3 \\
	ETH & 13 & 3 & 9 & 3 \\
	FJI & 4 & 1 & 8 & 1 \\
	FRA & 120 & 11 & 16 & 11 \\
	FSM & 1 & 0 & 0 & 0 \\
	GBR & 67 & 1 & 1 & 1 \\
	GEO & 10 & 2 & 12 & 2 \\
	GHA & 4 & 0 & 1 & 0 \\
	GIN & 1 & 0 & 1 & 0 \\
	GLP & 0 & 0 & 1 & 0 \\
	GMB & 1 & 0 & 0 & 0 \\
	GRC & 61 & 7 & 11 & 7 \\
	GRD & 0 & 0 & 2 & 0 \\
	GRL & 8 & 0 & 0 & 0 \\
	GTM & 55 & 15 & 38 & 15 \\
	GUF & 0 & 0 & 1 & 0 \\
	GUM & 0 & 0 & 1 & 0 \\
	HKG & 3 & 0 & 0 & 0 \\
	HND & 26 & 6 & 14 & 6 \\
	HRV & 9 & 0 & 1 & 0 \\
	HTI & 48 & 6 & 24 & 8 \\
	HUN & 9 & 0 & 0 & 0 \\
	IDN & 136 & 54 & 160 & 71 \\
	IND & 173 & 39 & 87 & 43 \\
	IRL & 2 & 0 & 0 & 0 \\
	IRN & 10 & 2 & 9 & 2 \\
	IRQ & 2 & 0 & 1 & 0 \\
	ISL & 4 & 0 & 0 & 0 \\
	ISR & 6 & 0 & 0 & 0 \\
	ITA & 314 & 27 & 40 & 30 \\
	JAM & 10 & 4 & 6 & 4 \\
	JOR & 1 & 0 & 0 & 0 \\
	JPN & 114 & 35 & 62 & 41 \\
	KAZ & 3 & 1 & 1 & 1 \\
	KEN & 18 & 3 & 17 & 3 \\
	KGZ & 8 & 3 & 14 & 4 \\
	KHM & 4 & 0 & 1 & 0 \\
	KIR & 1 & 0 & 0 & 0 \\
	KOR & 16 & 6 & 16 & 7 \\
	LAO & 4 & 0 & 4 & 0 \\
	LBN & 1 & 0 & 0 & 0 \\
	LBR & 1 & 0 & 0 & 0 \\
	LBY & 1 & 1 & 1 & 1 \\
	LCA & 0 & 0 & 2 & 0 \\
	LIE & 2 & 0 & 0 & 0 \\
	LKA & 33 & 10 & 33 & 11 \\
	LSO & 2 & 0 & 0 & 0 \\
	LUX & 15 & 0 & 1 & 0 \\
	LVA & 1 & 0 & 0 & 0 \\
	MAC & 0 & 0 & 1 & 0 \\
	MAR & 9 & 2 & 4 & 2 \\
	MDA & 2 & 0 & 0 & 0 \\
	MDG & 16 & 1 & 11 & 1 \\
	MEX & 61 & 18 & 53 & 28 \\
	MKD & 1 & 1 & 5 & 1 \\
	MLT & 1 & 0 & 0 & 0 \\
	MMR & 28 & 7 & 22 & 7 \\
	MNE & 4 & 0 & 0 & 0 \\
	MNG & 1 & 0 & 2 & 0 \\
	MOZ & 6 & 1 & 2 & 1 \\
	MRT & 1 & 0 & 0 & 0 \\
	MTQ & 1 & 0 & 2 & 0 \\
	MUS & 1 & 0 & 0 & 0 \\
	MWI & 5 & 1 & 3 & 1 \\
	MYS & 17 & 3 & 13 & 3 \\
	NER & 0 & 0 & 1 & 0 \\
	NGA & 2 & 0 & 3 & 0 \\
	NIC & 38 & 2 & 9 & 2 \\
	NLD & 6 & 0 & 0 & 0 \\
	NOR & 80 & 2 & 2 & 2 \\
	NPL & 142 & 18 & 45 & 25 \\
	NZL & 46 & 8 & 10 & 8 \\
	OMN & 0 & 0 & 2 & 0 \\
	PAK & 89 & 19 & 46 & 21 \\
	PAN & 16 & 2 & 17 & 2 \\
	PER & 103 & 13 & 42 & 13 \\
	PHL & 153 & 62 & 138 & 75 \\
	PNG & 20 & 5 & 22 & 5 \\
	POL & 23 & 0 & 0 & 0 \\
	PRI & 2 & 0 & 5 & 0 \\
	PRK & 13 & 0 & 7 & 0 \\
	PRT & 30 & 5 & 7 & 5 \\
	PRY & 3 & 0 & 2 & 0 \\
	REU & 0 & 0 & 2 & 0 \\
	ROU & 20 & 2 & 9 & 2 \\
	RUS & 53 & 2 & 11 & 2 \\
	RWA & 11 & 3 & 18 & 3 \\
	SAU & 2 & 0 & 1 & 0 \\
	SCG & 0 & 0 & 1 & 0 \\
	SLB & 3 & 2 & 5 & 2 \\
	SLE & 18 & 2 & 6 & 2 \\
	SLV & 32 & 8 & 17 & 10 \\
	SOM & 3 & 0 & 1 & 0 \\
	SPI & 0 & 0 & 1 & 0 \\
	SRB & 11 & 1 & 6 & 1 \\
	STP & 0 & 0 & 1 & 0 \\
	SUR & 1 & 0 & 0 & 0 \\
	SVK & 4 & 0 & 1 & 0 \\
	SVN & 13 & 2 & 4 & 2 \\
	SWE & 8 & 0 & 0 & 0 \\
	SYC & 2 & 0 & 1 & 0 \\
	SYR & 2 & 1 & 1 & 1 \\
	TCD & 0 & 0 & 1 & 0 \\
	THA & 18 & 2 & 17 & 2 \\
	TJK & 6 & 2 & 30 & 2 \\
	TLS & 1 & 1 & 2 & 1 \\
	TON & 1 & 0 & 0 & 0 \\
	TTO & 1 & 1 & 4 & 1 \\
	TUN & 4 & 0 & 0 & 0 \\
	TUR & 43 & 6 & 14 & 6 \\
	TWN & 47 & 12 & 18 & 13 \\
	TZA & 15 & 3 & 7 & 3 \\
	UGA & 36 & 9 & 22 & 9 \\
	UKR & 4 & 0 & 0 & 0 \\
	URY & 3 & 0 & 1 & 0 \\
	USA & 191 & 16 & 38 & 17 \\
	UZB & 0 & 0 & 2 & 0 \\
	VAT & 1 & 0 & 0 & 0 \\
	VCT & 2 & 0 & 3 & 0 \\
	VEN & 33 & 4 & 10 & 5 \\
	VIR & 0 & 0 & 1 & 0 \\
	VNM & 54 & 20 & 56 & 21 \\
	VUT & 5 & 1 & 6 & 1 \\
	WSM & 3 & 0 & 0 & 0 \\
	YEM & 8 & 3 & 6 & 3 \\
	ZAF & 28 & 2 & 4 & 2 \\
	ZMB & 2 & 0 & 1 & 0 \\
	ZWE & 4 & 0 & 1 & 0 \\
\end{supertabular}


\end{document}